\documentclass[12pt,A4]{article}
\usepackage[margin=1in,footskip=0.25in]{geometry}

\usepackage[comma]{natbib}
\usepackage{algorithm}
\usepackage{algorithmic}
\usepackage{graphicx}
\usepackage{booktabs} 
\usepackage{bm}
\usepackage{amsmath, amsfonts}
\usepackage{mathtools}
\usepackage{afterpage}
\usepackage{multirow}
\usepackage{subfig}
\usepackage{caption} 
\def\x{\bm{x}}   
    
\def\a{\bm{a}} \def\b{\bm{b}} 
\def\h{\bm{h}}
\def\W{\bm{W}} \def\I{\bm{I}} 
\def\S{\bm{S}}
\def\B{\bm{B}}
\def\C{\bm{C}}

\def \RRR {\mathbb{R}} \def \EEE {\mathbb{E}}

\newtheorem{lemma}{Lemma}
\newtheorem{remark}{Remark}

\DeclarePairedDelimiter{\diagfences}{(}{)}
\newcommand{\diag}{\operatorname{diag}\diagfences}

\begin{document}
	
	\title{An Effective and Efficient Initialization Scheme for Training Multi-layer Feedforward Neural Networks}

	\author{Zebin Yang$^1$$^*$, Hengtao Zhang$^1$$^*$, Agus Sudjianto$^2$, and Aijun Zhang$^1$$^{**}$\\
		{\normalsize  $^1$Department of Statistics and Actuarial Science, The University of Hong Kong,}\\
		{\normalsize Pokfulam Road, Hong Kong}\\
		{\normalsize $^2$Corporate Model Risk, Wells Fargo, USA}}
	
	\date{}
	\maketitle
	\def\thefootnote{*}\footnotetext{These authors contributed equally to this work}\def\thefootnote{\arabic{footnote}}
	\def\thefootnote{**}\footnotetext{Corresponding author: ajzhang@hku.hk}\def\thefootnote{\arabic{footnote}}

	\begin{abstract}
	Network initialization is the first and critical step for training neural networks. In this paper, we propose a novel network initialization scheme based on the celebrated Stein's identity. By viewing multi-layer feedforward neural networks as cascades of multi-index models, the projection weights to the first hidden layer are initialized using eigenvectors of the cross-moment matrix between the input's second-order score function and the response. The input data is then forward propagated to the next layer and such a procedure can be repeated until all the hidden layers are initialized. Finally, the weights for the output layer are initialized by generalized linear modeling. Such a proposed SteinGLM method is shown through extensive numerical results to be much faster and more accurate than other popular methods commonly used for training neural networks.
	\end{abstract}

	\section{Introduction} \label{Introduction}
Neural networks have shown extraordinary prediction performance in numerous learning tasks, while training a multi-layer neural network from scratch is not that easy. It may suffer from slow convergence, local optimum, overfitting, and many other problems. Accordingly, researchers have tried to solve these issues from different perspectives, including novel network architectures (e.g., the residual network;~\citealp{he2016deep}) and new optimizers (e.g., the Adam optimizer;~\citealp{kingma2014adam}). Recently, it is found that network initialization is also a critical step, and a good initialization scheme can help achieve almost-perfect performance~\citep{mishkin2016}.  

Multi-layer feedforward neural networks can be viewed as cascades of multi-index models due to its hierarchical architecture~\citep{yang2017high}. Given a dataset $\{\x_{i},y_{i}\}_{i\in[n]}$ with $\bm{x} \in \mathbb{R}^{d}$ being the input variables and $y$ being the response, a multi-index model is formulated as follows,
\begin{equation}\label{MIM}
\begin{gathered}
\EEE(y|\x) = g\left(\left\langle \x, \boldsymbol{\beta}_{1}\right\rangle, \ldots,\left\langle \x, \boldsymbol{\beta}_{k}\right\rangle\right),
\end{gathered}
\end{equation}
where $\left\{\boldsymbol{\beta}_{j}\right\}_{j \in[k]} \subseteq \mathbb{R}^{d}$ are the projection indices of interest and $g$ is an unknown function that maps the $k$ projected data to the response. For the sake of identifiability, the projection indices are usually assumed to be mutually orthogonal with unit norm. It is closely related to a feedforward neural network with $k$ nodes in the first hidden layer. The difference lies in the estimation of the link function $g$ such that the feedforward neural network estimates $g$ by multi-layer parametric transformation, while the multi-index model estimates $g$ in a nonparametric way.

The connection between multi-layer feedforward neural networks and multi-index models makes it feasible to quickly estimate the hidden layer weights without knowing the link function. For example,~\cite{yang2017learning} proposed to use Stein's identity~\citep{stein2004use} for estimating $\left\{\boldsymbol{\beta}_{j}\right\}_{j \in[k]}$ in (\ref{MIM}) under the subspace sparsity assumption. Similar methods based on Stein's identity have been employed for the estimation of single-index models~\citep{yang2017high}, additive index models~\citep{balasubramanian2018tensor}, index volatility models~\citep{na2018high}, and varying index coefficient models~\citep{na2019high}. Moreover, Stein's identity has also been utilized to estimate a single hidden layer network~\citep{janzamin2015beating, sedghi2014provable} with guaranteed risk bounds in high probability. 	

In this paper, we propose a novel network initialization method (SteinGLM) based on Stein's identity and generalized linear models (GLM). For the first hidden layer, the weight matrix can be quickly initialized using the second-order Stein's identity: a) we first calculate the empirical cross-moment matrix between the input's second-order score function and the response; b) the eigenvalue decomposition is performed on the cross-moment matrix; and c) the weight matrix initialization is obtained by collecting the top unit norm eigenvectors multiplied by a constant scaling factor. The same procedure is sequentially utilized to initialize the remaining higher hidden layers. Finally, the output layer weights and biases are initialized using GLM subject to proper regularizations.

\subsection{Related Work}
Over the past few decades, a lot of network initialization methods have been proposed to overcome the above-mentioned training difficulties. In general, these methods can be roughly divided into two categories based on whether the information of the response is explicitly considered or not.

When the response variables are given, a multi-layer feedforward neural network can be initialized via least squares solutions~\citep{martens1996stochastically}. As all the hidden layers are randomly initialized, it was suggested to forward propagate the input variables to the last hidden layer, and then the output layer weights can be estimated via least squares solutions~\citep{yam1997extended, yam1997new, cho1999training, yam2000weight, yam2001feedforward}.  Alternatively,  since the output layer weights are estimated, we can estimate the expected output of the previous hidden layer. Such a procedure can be further used for initializing all the hidden layers~\citep{erdogmus2005linear}.

Another direct approach is to build a surrogate model between the input variables and the response. For example, a pruning-based neural network is proposed for establishing a connection from decision trees to compressed single hidden layer networks~\citep{setiono1999mapping}. In~\cite{biau2019neural}, the authors proposed a neural random forest model for initializing a two-hidden-layer network. Other than tree-based methods, partial least squares regression can also be employed to initialize a single hidden layer network with identity activations~\citep{hsiao2003partial}. However, these methods focus on shallow networks only. A recent work that maps decision trees to deep neural networks can be referred to as~\citet{humbird2018deep}, where a lot of decision trees are first fitted on the data and then the neural network will be initialized according to the fitted trees.

The second type of approach prevents the forward/backward backpropagation of signals/gradients from exploding or vanishing, such that the information can freely flow through the hidden layers.  It is known that appropriately scaled random weights with Gaussian or uniform distributions can help achieve this goal, and the optimal scaling factor may vary when different activations are used~\citep{kumar2017weight}. For example, the Glorot initialization is shown to be the best for identity activations~\citep{glorot2010understanding}. The He initialization is designed for ReLU activations~\citep{he2015delving}. Extensive literature has been devoted to the selection of random generators. As there is no distributional assumption about the input variables, the connecting weights can be generated following uniform distribution and the appropriate interval parameters can be obtained by solving a linear interval tolerance problem~\citep{adam2014solving}. Another recent work analyzed the change of gradient magnitudes from the output layer to the first hidden layer and developed a random walk framework for determining the optimal scaling factors for different activations~\citep{sussillo2014random}. 

Orthogonal initialization is an alternative approach to Gaussian and uniform random initialization. When the activation is identity, the product of multiple orthogonal weight matrices is still an orthogonal matrix with all of its singular values equal to one. In contrast, the singular values of a Gaussian random matrix would spread, and most of them would shrink to zero as the number of hidden layers is large~\citep{saxe2013exact}. Consequently, only a few directions can be updated during the process of error backpropagation and the truly effective direction may be ignored. It should be noted that a similar conclusion can be derived when the activation functions are nonlinear, while the appropriate scaling factor may vary for different activations. The efficiency of orthogonal initialization can also be verified from the perspective of dynamical isometry, where superior performance is observed in speeding up the training performance~\citep{pennington2017resurrecting, pennington2018emergence, xiao2018dynamical}.

\subsection{Main Contributions}
In this paper, a novel network initialization method is proposed and to the best of our knowledge, this is the first work that employs Stein's identity for initializing multi-layer feedforward neural networks. The SteinGLM extends the aforementioned studies and its main advantages are summarized as follows,
\begin{itemize}
	\item All the parameters are initialized according to the information of the input variables and the response. Intuitively, our approach will have better performance than the purely random initializations;
	
	\item The weight matrices are initialized to be eigenvectors in the proposed SteinGLM method. They are inherently orthogonal and therefore share the benefits of orthogonal initialization; 
	
	\item By conducting extensive numerical experiments based on real datasets, the proposed SteinGLM method is shown to be more effective and efficient than other popular initialization benchmarks. A detailed ablation study is included for the drill-down analysis of the superior performance of the proposed method. Moreover, a specific strategy for applying the proposed SteinGLM method for initializing convolutional neural networks is provided.
\end{itemize}

The rest of the paper is organized as follows. Section~\ref{Method} introduces the background of Stein's identity and the proposed SteinGLM method. Numerical experiments are reported in Section~\ref{Experiments}. Finally, we conclude in Section~\ref{Conclusion} and discuss future research directions.

\section{Proposed Method} \label{Method}
A  multi-layer feedforward neural network is assumed to have one input layer, $L$ hidden layers and one output layer, where the $l$-th hidden layer has $N_{l}$ neurons. The transformation of data is defined as 
\begin{equation}
\begin{gathered}
\a_{l} = \W_{l}^{T} \h_{l-1}+ \b_{l}, \\
\h_{l} = f^{(h)} \left(\a_{l}\right),
\end{gathered}
\end{equation}
for $l = 1, 2, \cdots, L$. The symbols $\a_{l}, \h_{l}$ denote the values of the $l$-th hidden layer before and after the transformation, respectively. Note the input layer can be indexed by 0, such that $\h_{0} = \x$ and $N_{0} = d$. The weight matrix and  bias vector of the $l$-th hidden layer are denoted by $\W_{l}$ and $\b_{l}$. Finally, the output layer is given by
\begin{equation}
\begin{gathered}
\hat{y} = f^{(o)}\left(\W_{o}^{T} \h_{L} + \b_{o}\right), \\
\end{gathered}
\end{equation}
where $\W_{o}, \b_{o}$ are the parameters of the output layer. The output layer activation function is denoted by $f^{(o)}$, which may vary according to different learning tasks. We focus on the initialization of weights and biases of all the hidden layers $(\W_{1},\b_{1}), \ldots, (\W_{L},\b_{L})$ and the output layer $\W_{o}, \b_{o}$. For simplicity in the initialization step, the input variables are assumed to follow the standard normal distribution $\x \sim N\left(\bm{0}, \I_{d}\right)$.

\subsection{Stein's Identity}
Suppose the input variable $\x \in \RRR^{d}$ has a joint probability density $p(\x):\mathbb{R}^{d} \rightarrow \mathbb{R}$. The first-order Stein's identity is a celebrated lemma about the first-order score function~\citep{stein2004use}. 

\begin{lemma} \label{First_stein} (First-order Stein's Identity). Assume the density of $\x$ is differentiable and the first-order score function $\S_{1}(\x)=-\nabla_{\x} p(\x) / p(\x)$ exists. For any differentiable function $g: \mathbb{R}^{d} \rightarrow \mathbb{R} $ such that $\mathbb{E}[\nabla_{\x} g(\x)]$ exists and all the entries of $p(\x) g(\x)$ go to zero on the boundaries of support of $p(\x)$, we have
	$$
	\mathbb{E}\left[g(\x)\S_{1}(\x)\right]=\mathbb{E}\left[\nabla_{\x} g(\x)\right].
	$$
\end{lemma}

The first-order Stein's identity can be used to extract the weight vector of multi-index models defined in (\ref{MIM}). As $k=1$, it reduces to the single-index model and we have 
\begin{equation}\label{first_stein_sim}
\begin{split}
\mathbb{E}\left[g(\x)\S_{1}(\x)\right]=\mathbb{E}\left[g^{\prime}\left(\left\langle \x, \boldsymbol{\beta}_{1}\right\rangle\right)\right] \boldsymbol{\beta}_{1},
\end{split}
\end{equation}
where the expectation of the first order derivative is a constant term. Given $\mathbb{E}\left[g^{\prime}\left(\left\langle \x, \boldsymbol{\beta}_{1}\right\rangle\right)\right] \neq 0$, the vector $\boldsymbol{\beta}_{1}$ can be easily estimated, and such an estimator is easy to compute with good statistical properties~\citep{yang2017high}. 

Nevertheless, the first-order Stein's identity is not sufficient to estimate all the weight vectors when $k>1$. In this case, the second-order Stein's identity contains more information about the data and is shown to be much more robust for estimating the multi-index model~\citep{yang2017learning}. With reference to~\citep{janzamin2014score}, 
the higher-order score functions  can be iteratively defined by 
\begin{equation}\label{high_score_funcitons}
\begin{split}
\S_{m}(\x)=(-1)^{m} \frac{\nabla^{(m)} p(\x)}{p(\x)}, \ m\geq 1.
\end{split}
\end{equation}
We have the lemma of second-order Stein's identity. 
\begin{lemma} \label{Second_stein} (Second-order Stein's Identity). Assume the density of $\x$ is twice differentiable and the second-order score function $\S_{2}(\x)=\nabla^{2}_{\x} p(\x) / p(\x)$ exists. For any twice differentiable function $g:\RRR^{d} \rightarrow \RRR^{d \times d}$ such that $\mathbb{E}\left[\nabla_{\x}^{2} g(\x)\right]$ exists and all the entries of $g(\x) \nabla_{\x} p(\x)$ and $\nabla_{\x} g(\x) p(\x)$ go to zero on the boundaries of support of $p(\x)$, we have
	$$
	\mathbb{E}[g(\x) \S_{2}(\x)]=\mathbb{E}\left[\nabla^{2}_{\x} g(\x)\right].
	$$
\end{lemma}

Note that $\S_{1}(\x)$ is a vector, while $\S_{2}(\x)$ is a matrix. Provided that the condition  $\mathbb{E}\left[g^{\prime\prime}\left(\left\langle \x, \B\right\rangle\right)\right] \neq 0$, we can extract all the weight vectors $\B = [\boldsymbol{\beta}_{1}, \ldots, \boldsymbol{\beta}_{k}]$ in (\ref{MIM}) without knowing the explicit form of the link function $g$,
\begin{equation}\label{second_stein_mim}
\begin{split}
\mathbb{E}[y \S_{2}(\x)] = \B \C \B^{T} = \sum_{j=1}^{k} c_{j}\boldsymbol{\beta}_{j}\boldsymbol{\beta}_{j}^T,
\end{split}
\end{equation}
where $\C = \diag{c_{1},\ldots,c_{k}}$ denotes the expectation of the second-order derivative of the link function $g$, which is assumed to be non-zero. As $\B$ is assumed to be an orthonormal matrix, a direct estimator for $\B$ is to perform eigenvalue decomposition on $\bm{\Sigma} = \mathbb{E}[y \S_{2}(\x)]$, and the top $k$ unit norm eigenvectors can be collected as the estimator of $\B$. In practice, the cross-moment matrix $\bm{\Sigma} $ can be estimated empirically by
\begin{equation}\label{emp_sigma}
\begin{split}
\hat{\bm{\Sigma}}=\frac{1}{n} \sum_{i=1}^{n} y_{i} \S_{2}\left(\x_{i}\right),
\end{split}
\end{equation}
which requires the evaluation of the second-order score function for each data point. 

It is worthy to mention that there exist higher-order  Stein's identities. For example, the third-order score function is a tensor, and it contains more information about the data. The third-order Stein's identity has been employed to estimate additive-index models~\citep{balasubramanian2018tensor} and single hidden layer networks~\citep{janzamin2015beating}. However, the tensor operation is an expensive procedure. To calculate the score function requires about $O(nd^3)$ computational complexity and corresponding memory complexity, which is not an efficient choice in practice. Moreover, the third-order Stein's identity seems not to be suitable for initializing multi-layer feedforward neural networks. That is, the number of unique weight vectors generated by tensor decomposition may be different from the actual size of neurons, which adds uncertainty for the practical application. Therefore, in our proposed initialization method, it is recommended to utilize the second-order Stein's identity. 

\subsection{SteinGLM Algorithm}
\begin{figure}[!t]
	\vskip 0.2in
	\begin{center}
		\centerline{\includegraphics[width=1.0\columnwidth]{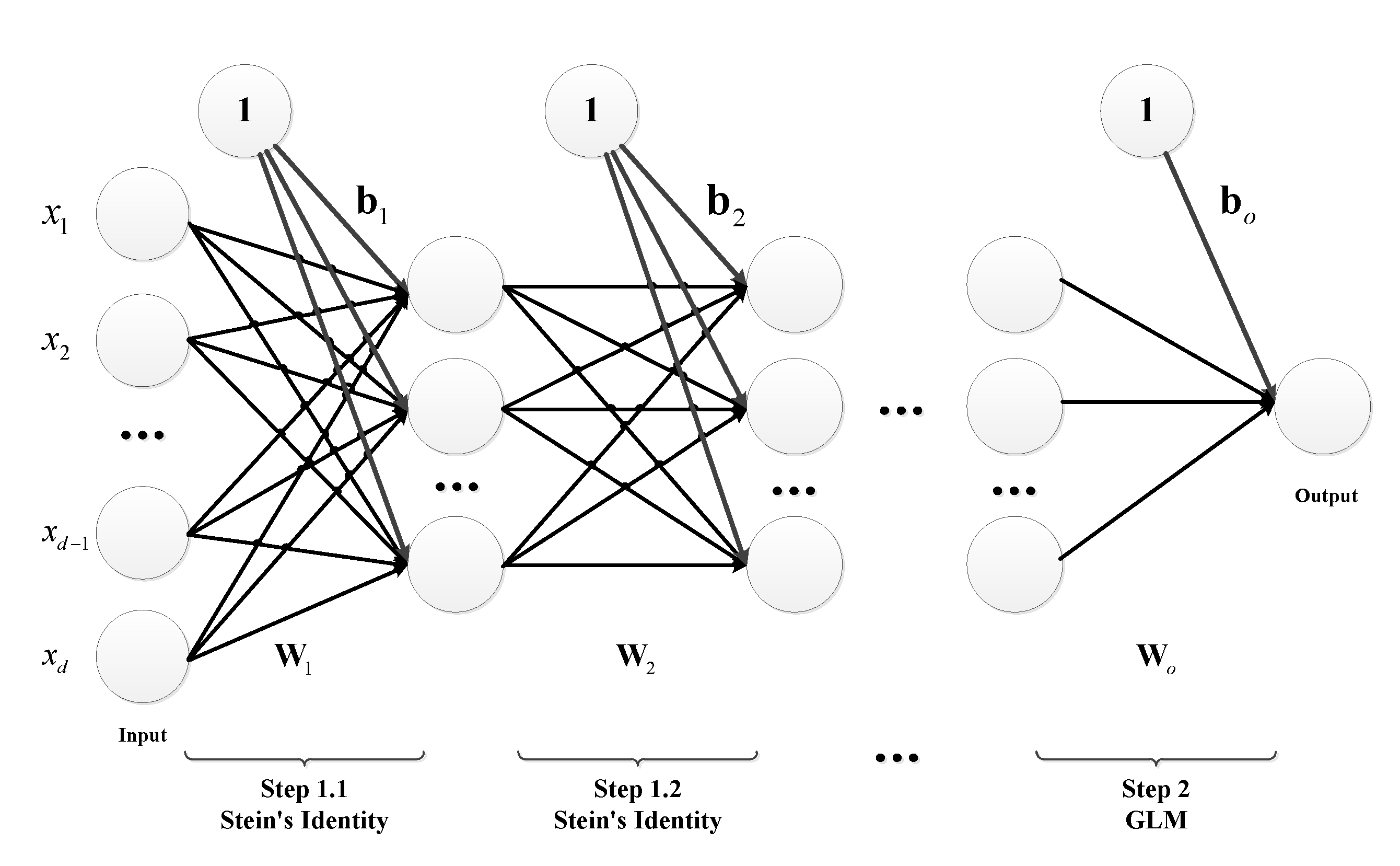}}
		\caption{The framework of the proposed SteinGLM method.}
		\label{framework}
	\end{center}
	\vskip -0.2in
\end{figure}

The proposed SteinGLM method performs a two-step procedure to initialize a multi-layer feedforward neural network from the first hidden layer to the output layer. The corresponding framework is visualized in Figure~\ref{framework}.

\textbf{1) Hidden Layer Initialization via Stein's identity}.  

Multi-layer neural networks can be viewed as the cascades of multi-index models defined in (\ref{MIM}). The weight matrix in the first hidden layer corresponds to the projection indices and the rest layers represent the unknown link function. 

Under the assumption $\x \sim N\left(\bm{0}, \I_{d}\right)$, the first-order and second-order score functions have the neat explicit forms $\S_{1}(\x) = \x$ and $\S_{2}(\x)=\x \x^{T}-\I_{d}$. The second-order Stein's identity enables us to initialize the first hidden layer weights of a multi-layer feedforward neural network via 
\begin{equation}\label{W_init}
\begin{split}
\W_{1} = \alpha \hat{\B},
\end{split}
\end{equation}
where $\hat{\B}$ is the top-$k$ eigenvectors of $\hat{\bm{\Sigma}}$ and $\alpha$ is a scaling factor for controlling the magnitude of weights. The bias vector can be accordingly specified to be
\begin{equation}\label{b_init}
\begin{split}
\b_{1} = -\frac{1}{n} \sum_{i=1}^{n} \W_{1}^{T}\x_{i},
\end{split}
\end{equation}
such that the empirical mean of $\a_{1} $ is initialized to be zero. Since zero is the most active point for the majority of activation functions, this calibration helps prevent the activations from being dead. 

Note that the biases are commonly initialized to be zero in the literature. However, such zero initialization is not suitable for activations whose expectation is non-zero, e.g., the sigmoid function. Without calibration, the initial values of $\a_{l}$ may deviate from the most active regions of the activation function. Consequently, the standard deviation of $\a_{l}$ will quickly shrinkage to zero with the increase of $l$, and thus the initialization will become invalid. Therefore, the bias calibration is a necessary procedure, and it can be verified through numerical study in the next section.  

The scaling factor $\alpha$ is introduced to offset the variance instability caused by  activation function during the first forward propagation. The optimal value of $\alpha$ varies for different activation functions, and more detailed discussion for choosing the proper scaling factor can be referred to as~\citet{kumar2017weight}. As for the proposed Stein's initialization scheme, the output of each hidden layer (given continuous activation functions) can be approximated using Taylor expansion, as follows,
\begin{equation} \label{taylor_expan}
f^{(h)}(x) \approx f^{(h)}(0) + \left(x - 0\right) \nabla{f^{(h)}}(0),
\end{equation}	
as $f^{(h)}$ is differentiable at 0. More precisely, we provide two examples that are commonly used in practice.
\begin{itemize}
	\item Hyperbolic tangent: $f^{(h)}(0) = 0, \nabla{f^{(h)}}(0) = 1$. We suggest to use a scaling factor $\alpha = 1$.
	\item Sigmoid: $f^{(h)}(0) = 0.5, \nabla{f^{(h)}}(0) = 0.25$. We can multiply $\hat{\B}$ with a scaling factor $\alpha = 4$.
\end{itemize}

\begin{remark}
	For both hyperbolic tangent and sigmoid, the maximal derivative is at the zero-point. Therefore, the scaling factors calculated by Taylor approximation are underestimated. In practice, slightly larger scaling factors can be used as compared to the suggested ones. 
\end{remark}

The above procedure can be successively performed to initialize the remaining hidden layers. Specifically, the output $\h_{1}$ from the first hidden layer can be viewed as a ``new'' input layer, and the weights and biases of the second hidden layer can be initialized using the second-order Stein's identity.  Repeat this procedure until all the hidden layers are initialized.

\begin{remark}
	In practice, the density of the input variables is usually unknown. There exit efficient estimation methods for score functions via score matching~\citep{hyvarinen2005estimation, lyu2009interpretation}, where the density function is not required. For our proposed SteinGLM method, score matching algorithms can be employed to better approximate the score function for each layer. As a trade-off, it would require a lot more computation and sacrifice efficiency. Thus in the present paper, we take the normal assumption and aim to develop a much faster yet accurate initialization algorithm.  
\end{remark}

\textbf{2) Output Layer Initialization via GLM}.

After all the hidden layers are initialized, the input variables can be forward propagated to the last hidden layer with output $\h_L$. Then, the final output layer can be initialized through generalized linear modeling (GLM), by treating $\h_{L}$ as predictors and $y$ as the response. The specific type of GLM can be determined by the learning tasks. For example, the mean squared error is employed for regression tasks
\begin{equation} \label{LossLS}
\min \frac{1}{n}\sum_{i=1}^{n} (y_i - \hat{y}_i)^2,
\end{equation}	
while the cross entropy loss is employed for logistic regression with classification tasks
\begin{equation} \label{LossCE}
\min -\frac{1}{n}\sum_{i=1}^{n} y_i\log\hat{y}_i + (1-y_i)\log(1-\hat{y}_i).
\end{equation}	
In practice, since the last hidden layer output may be poorly conditioned,  we also introduce the $\ell_{2}$-regularization to prevent the potential multicollinearity problem. The optimal regularization strength can be tuned by the simple grid search method. 

In summary, the proposed SteinGLM method is presented in Algorithm~\ref{Algo}. 
Its computation complexity is analyzed as follows. For the $(l+1)$-th hidden layer ($l=0,1,\cdots,L-1$), the second-order cross-moment, eigenvalue decomposition, and forward propagation requires $O(nN_{l}^2)$, $O(N_{l}^3)$, and $O(nN_{l}N_{l+1})$, respectively. The time complexity of GLM varies according to the learning tasks. For the example of ridge regression, it requires $\max{\{O(nN_{L}^2), O(N_{L}^3)\}}$. As $n \gg N_{L}$, asymptotically $O(nN_{L}^2)$ takes the dominant part; otherwise $O(N_{L}^3)$ will be the dominant, which is close to that of orthogonal initialization.

\begin{algorithm}[!t] 
	\caption{SteinGLM Algorithm} \label{Algo}
	\begin{algorithmic}
		\STATE {\bfseries Input:} Data $\{\x_{i}, y_{i}\}_{i \in [n]}$, activation functions $f^{(h)}, f^{(o)}$, hidden layer number $L$, number of neurons $\{N_{l}\}_{l \in [L]}$.
		\STATE Normalize the input variables and set $\h_{0} = \x$.
		\FOR{$l=1$ {\bfseries to} $L$}
		\STATE Compute the second-order score function for $\h_{l}$ and the corresponding cross-moment matrix $\hat{\bm{\Sigma}}$ via (\ref{emp_sigma});
		\STATE Perform eigenvalue decomposition to $\hat{\bm{\Sigma}}$, and assign $\W_{l}$ with the top $N_{l}$ eigenvectors;
		\STATE Assign $\b_{l} = -\frac{1}{n} \sum \W_{l}^{T}\h_{l-1}$;
		\STATE Forward propagate the data to the next layer.
		\ENDFOR
		\STATE Initialize $\W_{o}, \b_{o}$ for the output layer via GLM.
		\STATE {\bfseries Output:} The initialized network.
	\end{algorithmic}
\end{algorithm}

\section{Experiments} \label{Experiments}
In this section, we perform extensive numerical studies to test the performance of the proposed SteinGLM method.

\subsection{Experimental Settings}
The proposed method is tested on 10 real-world datasets, including 5 regression datasets and 5 binary classification datasets. All the samples with missing values are removed, and a summary of the tested datasets is given in Table~\ref{data_info}. In specific, the California Housing dataset is fetched from~\textsl{scikit-learn} package while the rest 4 regression datasets and 5 classification datasets are all directly obtained from the UCI repository. Each dataset is split for training (80\%) and test (20\%).

\begin{table}[!t]
	\caption{Datasets information.}
	\label{data_info}
	\begin{center}
		\begin{tabular}{ccccc}
			\toprule
			Data &            Name            & Samples & Features &   Task Type    \\ \midrule
			D1    &          Abalone           &  4177   &    8     &   Regression   \\
			D2    &     California Housing     &  20640  &    8     &   Regression   \\
			D3    & Combined Cycle Power Plant &  9568   &    4     &   Regression   \\
			D4    &      Electrical Grid       &  10000  &    11    &   Regression   \\
			D5    &      Parkinsons Tele       &  5875   &    19    &   Regression   \\
			D6    &           Adult            &  30162  &    14    & Classification \\
			D7    &       Bank Marketing       &  45211  &    16    & Classification \\
			D8    &           Magic            &  19020  &    10    & Classification \\
			D9    &        Mammographic        &   830   &    5     & Classification \\
			D10   &          Spambase          &  4601   &    57    & Classification \\ \bottomrule
		\end{tabular}
	\end{center}
\end{table}

A small validation set is further split from the training set for monitoring its generalization performance. Finally, we evaluate the root mean squared error (RMSE) and the area under the curve (AUC) for regression and binary classification tasks, respectively. Note all the data is preprocessed before the experiment. We perform one-hot encoding for categorical features, all the input variables are standardized with zero mean and unit variance, and the response variable is linearly scaled within 0 and 1.

The proposed method is compared with three commonly used network initialization methods, as follows,
\begin{itemize}
	\item Glorot Normal~\citep{glorot2010understanding}: for the $l$-th hidden layer, weights are randomly generated from the truncated normal distribution with zero mean and variance $2/(N_{l} + N_{l+1})$; 
	\item He Normal~\citep{he2015delving}: for the $l$-th hidden layer, weights are randomly generated from the truncated normal distribution with zero mean and variance $2/N_{l}$;
	\item Orthogonal~\citep{saxe2013exact}: weights are generated by collecting the singular vectors of a randomly generated Gaussian matrix.
\end{itemize}

We consider to initialize a multi-layer feedforward neural network with 10, 20, 30, and 40 hidden layers with $\min{\{d, 20\}}$ neurons per layer, and $d$ is the number of features after one-hot encoding. The commonly used hyperbolic tangent function is employed as the activation. The maximal training epochs are set to 200, with a mini-batch size of 500 but not larger than 20\% of the training sample size. Adam optimizer is used with an initial learning rate of 0.001. To avoid overfitting, we record the network parameters that achieve the best validation performance during training. All the experimental codes are written in Python with neural networks implemented using~\textsl{TensorFlow 2.0} platform. All the experiments are conducted on a multi-core CPU server and each experiment is repeated 10 times with the averaged performance reported.

\subsection{Results}
\begin{figure}[!ht]
	\centering
	\includegraphics[width=0.95\textwidth]{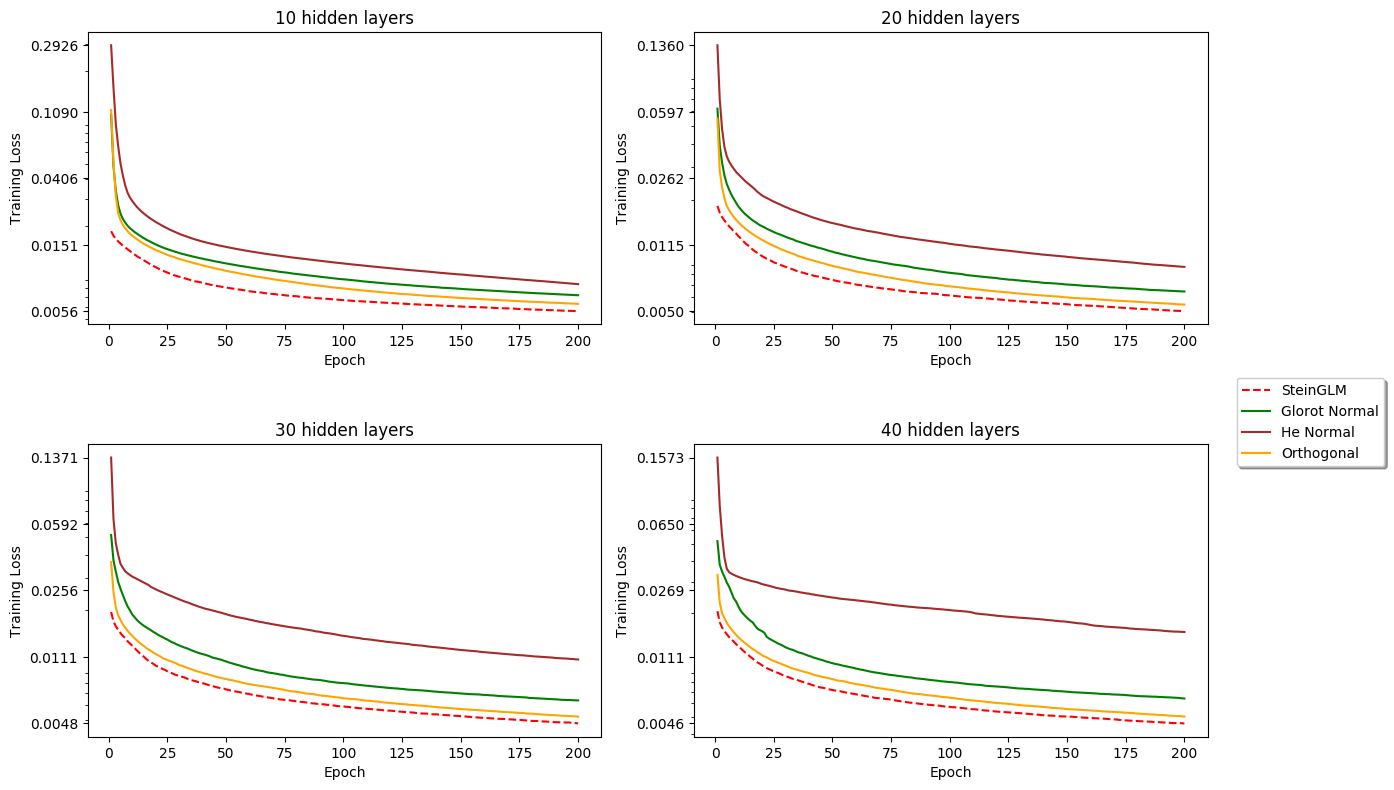}
	\caption{The trajectories of training losses averaged over all regression datasets. Note all the response variables are scaled within 0 and 1, the corresponding loss of each dataset are in a similar scale.}
	\label{traj_reg}
	
	\medskip
	\includegraphics[width=0.95\textwidth]{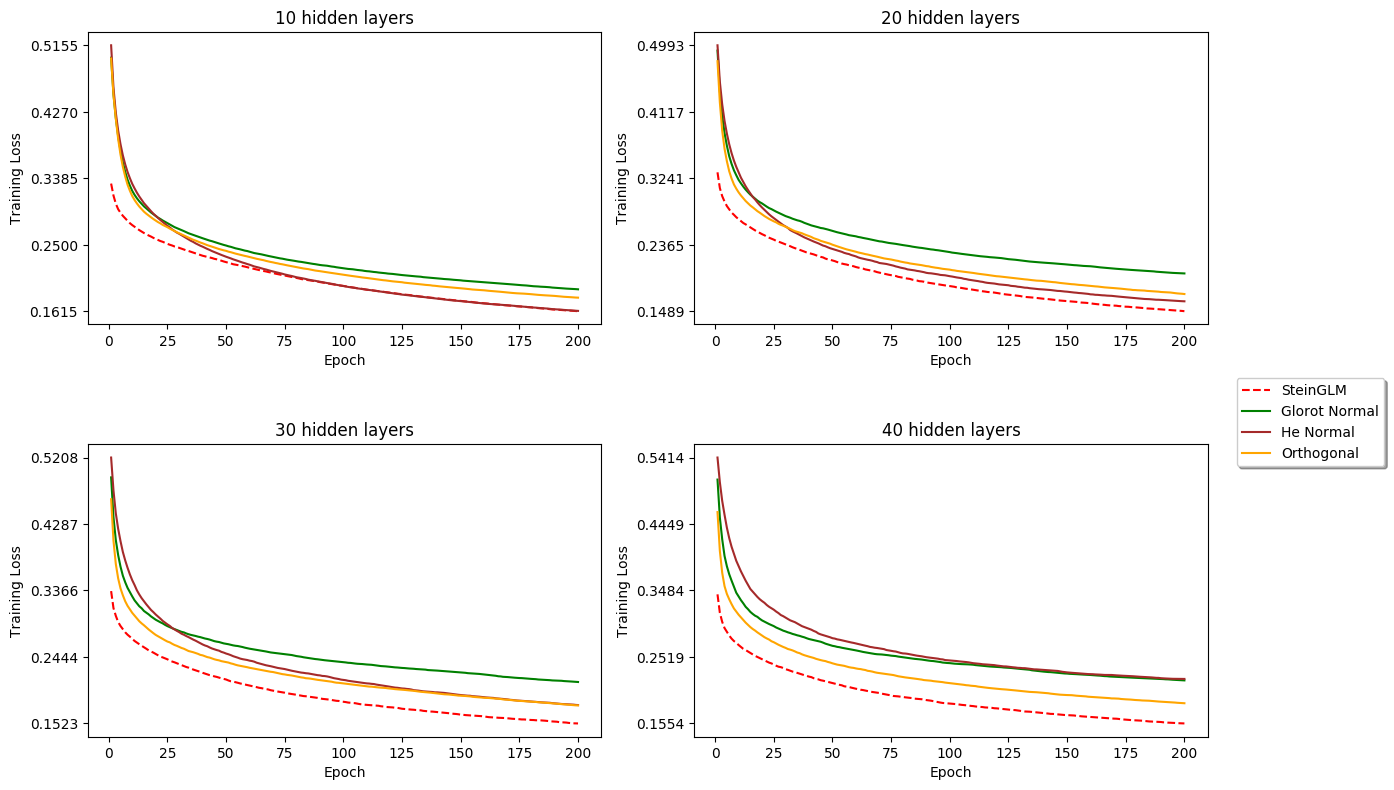}
	\caption{The trajectories of training losses averaged over all classification datasets. Note the cross-entropy loss of each each dataset are in a similar scale.}
	\label{traj_cls}
\end{figure}

The trajectories of training losses with different hidden layers are reported in Figures~\ref{traj_reg}--\ref{traj_cls}. For simplicity, the averaged training losses are used, and each point in the figures is averaged over the tested datasets and ten repetitions, and the y-axis is on a log scale for regression tasks. The experimental results show that the proposed SteinGLM method achieves superior performance in most of the datasets as compared with the commonly used initialization methods, i.e., Glorot Normal, He Normal, and Orthogonal. Networks initialized with SteinGLM converge much faster than that of the benchmarks. Moreover, it can be observed that the three benchmarks can easily get stuck into less optimal solutions.

The averaged test set RMSE and AUC for regression and binary classification tasks are reported in Table~\ref{reg_res} and Table~\ref{cls_res}, respectively. Both the average and standard deviation of test set performance are reported, where the best performing results are highlighted in bold. It can be observed that the proposed SteinGLM method performs the best among the compared benchmarks in both regression and classification datasets.

\begin{table}[!t]
	\caption{Test set RMSE of regression datasets.}
	\label{reg_res}
	\renewcommand\tabcolsep{3pt}
	\begin{center}
		\begin{tabular}{cccccc}
			\toprule
			Data               & Layers &           SteinGLM           &   Glorot Normal   &     He Normal     &          Orthogonal          \\ \midrule
			\multirow{4}{*}{D1}       &   10   & $\mathbf{0.0755}$$\pm$0.0024 & 0.0773$\pm$0.0024 & 0.0850$\pm$0.0038 &      0.0768$\pm$0.0027       \\
			&   20   & $\mathbf{0.0755}$$\pm$0.0025 & 0.0765$\pm$0.0024 & 0.0859$\pm$0.0062 &      0.0761$\pm$0.0025       \\
			&   30   & $\mathbf{0.0755}$$\pm$0.0025 & 0.0766$\pm$0.0028 & 0.0896$\pm$0.0069 &      0.0757$\pm$0.0027       \\
			&   40   & $\mathbf{0.0755}$$\pm$0.0025 & 0.0763$\pm$0.0021 & 0.0949$\pm$0.0036 &      0.0760$\pm$0.0022       \\ \midrule
			\multirow{4}{*}{D2} &   10   & $\mathbf{0.1090}$$\pm$0.0020 & 0.1150$\pm$0.0022 & 0.1154$\pm$0.0025 &      0.1110$\pm$0.0029       \\
			&   20   & $\mathbf{0.1085}$$\pm$0.0020 & 0.1138$\pm$0.0021 & 0.1171$\pm$0.0028 &      0.1093$\pm$0.0021       \\
			&   30   & $\mathbf{0.1084}$$\pm$0.0019 & 0.1136$\pm$0.0017 & 0.1222$\pm$0.0051 &      0.1090$\pm$0.0019       \\
			&   40   & $\mathbf{0.1089}$$\pm$0.0018 & 0.1146$\pm$0.0016 & 0.1283$\pm$0.0113 & $\mathbf{0.1089}$$\pm$0.0012 \\ \midrule
			\multirow{4}{*}{D3}        &   10   & $\mathbf{0.0543}$$\pm$0.0008 & 0.0554$\pm$0.0010 & 0.0586$\pm$0.0046 &      0.0555$\pm$0.0010       \\
			&   20   & $\mathbf{0.0542}$$\pm$0.0008 & 0.0551$\pm$0.0011 & 0.0600$\pm$0.0048 &      0.0554$\pm$0.0011       \\
			&   30   & $\mathbf{0.0542}$$\pm$0.0008 & 0.0551$\pm$0.0009 & 0.0722$\pm$0.0114 &      0.0553$\pm$0.0010       \\
			&   40   & $\mathbf{0.0543}$$\pm$0.0008 & 0.0550$\pm$0.0010 & 0.0976$\pm$0.0318 &      0.0550$\pm$0.0011       \\ \midrule
			\multirow{4}{*}{D4}   &   10   & $\mathbf{0.0354}$$\pm$0.0012 & 0.0484$\pm$0.0097 & 0.0932$\pm$0.0130 &      0.0403$\pm$0.0022       \\
			&   20   & $\mathbf{0.0338}$$\pm$0.0013 & 0.0410$\pm$0.0028 & 0.1076$\pm$0.0145 &      0.0369$\pm$0.0011       \\
			&   30   & $\mathbf{0.0333}$$\pm$0.0010 & 0.0393$\pm$0.0018 & 0.1181$\pm$0.0090 &      0.0364$\pm$0.0011       \\
			&   40   & $\mathbf{0.0333}$$\pm$0.0010 & 0.0399$\pm$0.0022 & 0.1373$\pm$0.0199 &      0.0354$\pm$0.0014       \\ \midrule
			\multirow{4}{*}{D5}   &   10   & $\mathbf{0.1152}$$\pm$0.0049 & 0.1258$\pm$0.0055 & 0.1427$\pm$0.0094 &      0.1197$\pm$0.0048       \\
			&   20   & $\mathbf{0.1089}$$\pm$0.0065 & 0.1212$\pm$0.0098 & 0.1687$\pm$0.0086 &      0.1156$\pm$0.0047       \\
			&   30   & $\mathbf{0.1094}$$\pm$0.0046 & 0.1222$\pm$0.0093 & 0.1753$\pm$0.0080 &      0.1140$\pm$0.0071       \\
			&   40   & $\mathbf{0.1100}$$\pm$0.0066 & 0.1201$\pm$0.0114 & 0.1796$\pm$0.0116 &      0.1124$\pm$0.0045       \\ \bottomrule
		\end{tabular}
	\end{center}
\end{table}

\begin{table}[!t]
	\caption{Test set AUC of binary classification datasets.}
	\label{cls_res}
	\renewcommand\tabcolsep{3pt}
	\begin{center}
		\begin{tabular}{cccccc}
			\toprule
			Data             & Layers &           SteinGLM           &   Glorot Normal   &    He  Normal     &    Orthogonal     \\ \midrule
			\multirow{4}{*}{D6}      &   10   & $\mathbf{0.9020}$$\pm$0.0033 & 0.8988$\pm$0.0032 & 0.8915$\pm$0.0023 & 0.9003$\pm$0.0031 \\
			&   20   & $\mathbf{0.9027}$$\pm$0.0033 & 0.8983$\pm$0.0033 & 0.8852$\pm$0.0036 & 0.9004$\pm$0.0039 \\
			&   30   & $\mathbf{0.9034}$$\pm$0.0035 & 0.8982$\pm$0.0029 & 0.8866$\pm$0.0046 & 0.8989$\pm$0.0032 \\
			&   40   & $\mathbf{0.9019}$$\pm$0.0028 & 0.8972$\pm$0.0028 & 0.8896$\pm$0.0044 & 0.8988$\pm$0.0031 \\ \midrule
			\multirow{4}{*}{D7} &   10   & $\mathbf{0.9220}$$\pm$0.0030 & 0.9172$\pm$0.0037 & 0.9046$\pm$0.0040 & 0.9197$\pm$0.0024 \\
			&   20   & $\mathbf{0.9226}$$\pm$0.0026 & 0.9149$\pm$0.0023 & 0.8979$\pm$0.0053 & 0.9176$\pm$0.0032 \\
			&   30   & $\mathbf{0.9222}$$\pm$0.0026 & 0.9159$\pm$0.0023 & 0.8990$\pm$0.0025 & 0.9166$\pm$0.0024 \\
			&   40   & $\mathbf{0.9211}$$\pm$0.0028 & 0.9137$\pm$0.0036 & 0.9008$\pm$0.0074 & 0.9156$\pm$0.0023 \\ \midrule
			\multirow{4}{*}{D8}      &   10   & $\mathbf{0.9314}$$\pm$0.0041 & 0.9285$\pm$0.0036 & 0.9218$\pm$0.0034 & 0.9302$\pm$0.0038 \\
			&   20   & $\mathbf{0.9302}$$\pm$0.0032 & 0.9254$\pm$0.0036 & 0.9121$\pm$0.0051 & 0.9286$\pm$0.0048 \\
			&   30   & $\mathbf{0.9295}$$\pm$0.0029 & 0.9264$\pm$0.0035 & 0.9105$\pm$0.0052 & 0.9272$\pm$0.0035 \\
			&   40   & $\mathbf{0.9280}$$\pm$0.0040 & 0.9251$\pm$0.0034 & 0.9125$\pm$0.0066 & 0.9262$\pm$0.0038 \\ \midrule
			\multirow{4}{*}{D9}  &   10   & $\mathbf{0.8871}$$\pm$0.0192 & 0.8828$\pm$0.0191 & 0.8711$\pm$0.0147 & 0.8818$\pm$0.0201 \\
			&   20   & $\mathbf{0.8852}$$\pm$0.0189 & 0.8649$\pm$0.0174 & 0.8438$\pm$0.0294 & 0.8742$\pm$0.0240 \\
			&   30   & $\mathbf{0.8832}$$\pm$0.0188 & 0.8523$\pm$0.0282 & 0.8333$\pm$0.0301 & 0.8780$\pm$0.0283 \\
			&   40   & $\mathbf{0.8825}$$\pm$0.0183 & 0.8570$\pm$0.0316 & 0.8326$\pm$0.0261 & 0.8591$\pm$0.0420 \\ \midrule
			\multirow{4}{*}{D10}     &   10   & $\mathbf{0.9787}$$\pm$0.0040 & 0.9730$\pm$0.0044 & 0.9680$\pm$0.0047 & 0.9757$\pm$0.0032 \\
			&   20   & $\mathbf{0.9798}$$\pm$0.0026 & 0.9677$\pm$0.0057 & 0.9514$\pm$0.0081 & 0.9720$\pm$0.0044 \\
			&   30   & $\mathbf{0.9793}$$\pm$0.0034 & 0.9703$\pm$0.0024 & 0.9407$\pm$0.0102 & 0.9717$\pm$0.0046 \\
			&   40   & $\mathbf{0.9779}$$\pm$0.0037 & 0.9666$\pm$0.0068 & 0.9403$\pm$0.0129 & 0.9710$\pm$0.0058 \\ \bottomrule
		\end{tabular}
	\end{center}
\end{table}

\subsection{Ablation Study}
To investigate what makes SteinGLM achieve superior performance, ablation studies are further conducted. That is, we first consider to include the GLM initialization (for the last layer) for each of the three benchmark methods. Similarly, we also consider SteinGLM without GLM initialization, where both hidden and output layers are initialized using the second-order Stein's method. In other words, we make a fair comparison among the Stein's method, Glorot Normal, He Normal, and Orthogonal, under two scenarios when GLM is used or not.

\begin{table}[!ht]
	\caption{Test set RMSE of regression datasets with or without GLM.}
	\label{ablation_reg_res}
	\begin{center}
		\begin{tabular}{cccccc}
			\toprule
			GLM          & Layers & He Normal & Glorot Normal &   Orthogonal    &    SteinGLM     \\ \midrule
			\multirow{4}{*}{No}  &   10   &  0.0990   &    0.0844     & \textbf{0.0807} &     0.0808      \\
			&   20   &  0.1078   &    0.0815     &     0.0787      & \textbf{0.0772} \\
			&   30   &  0.1155   &    0.0814     &     0.0781      & \textbf{0.0776} \\
			&   40   &  0.1276   &    0.0812     &     0.0775      & \textbf{0.0774} \\ \midrule
			\multirow{4}{*}{Yes} &   10   &  0.0847   &    0.0833     &     0.0788      & \textbf{0.0779} \\
			&   20   &  0.0995   &    0.1018     &     0.0774      & \textbf{0.0762} \\
			&   30   &  0.1106   &    0.1128     &     0.0771      & \textbf{0.0762} \\
			&   40   &  0.1148   &    0.1125     &     0.0772      & \textbf{0.0764} \\ \bottomrule
		\end{tabular}
	\end{center}
	
	\caption{Test set AUC of classification datasets with or without GLM.}
	\label{ablation_cls_res}
	\begin{center}
		\begin{tabular}{cccccc}
			\toprule
			GLM          & Layers & He Normal & Glorot Normal & Orthogonal & SteinGLM \\ \midrule
			\multirow{4}{*}{No}  &   10   &  0.9163   &    0.9245     &   0.9258   &  \textbf{0.9264}  \\
			&   20   &  0.9044   &    0.9196     &   0.9233   &  \textbf{0.9266}  \\
			&   30   &  0.9019   &    0.9164     &   0.9235   &  \textbf{0.9254}  \\
			&   40   &  0.9017   &    0.9169     &   0.9192   &  \textbf{0.9251}  \\ \midrule
			\multirow{4}{*}{Yes} &   10   &  0.9123   &    0.9218     &   0.9257   &  \textbf{0.9276}  \\
			&   20   &  0.9008   &    0.9207     &   0.9250   &  \textbf{0.9274}  \\
			&   30   &  0.9017   &    0.9176     &   0.9246   &  \textbf{0.9270}  \\
			&   40   &  0.9026   &    0.8554     &   0.9237   &  \textbf{0.9260}  \\ \bottomrule
		\end{tabular}
	\end{center}
\end{table}

As the test performance of each dataset is in a relatively similar scale, we report the average performance in Table~\ref{ablation_reg_res} and Table~\ref{ablation_cls_res}. The results confirm that the proposed Stein's method dominants the commonly used benchmark initialization methods. The only exception is the regression datasets with 10 hidden layers, where Orthogonal initialization without GLM achieves slightly superior performance than the proposed model. 

The contribution of GLM initialization can also be observed. In SteinGLM, the test performance gets improved in all the cases as GLM is used. However, the training process may get stuck in locally optimal solutions, e.g., when He Normal or Glorot Normal is used. Therefore, it can be concluded that both Stein's method and GLM are necessary and indispensable for the success of the proposed method.	

\subsection{SteinGLM for Convolutional Neural Networks}
Finally, we move one step further and extend the proposed SteinGLM method for initializing convolutional neural networks (CNNs). For a convolutional layer, image patches are convolved by filters (each filter is a matrix containing weight coefficients) to find matched patterns. Each filter will slide over the entire image, and output a feature map. A typical CNN stacks multiple convolutional layers, such that the important patterns can be automatically extracted.

\begin{figure}[!t]
	\centering
	\includegraphics[width=0.75\textwidth]{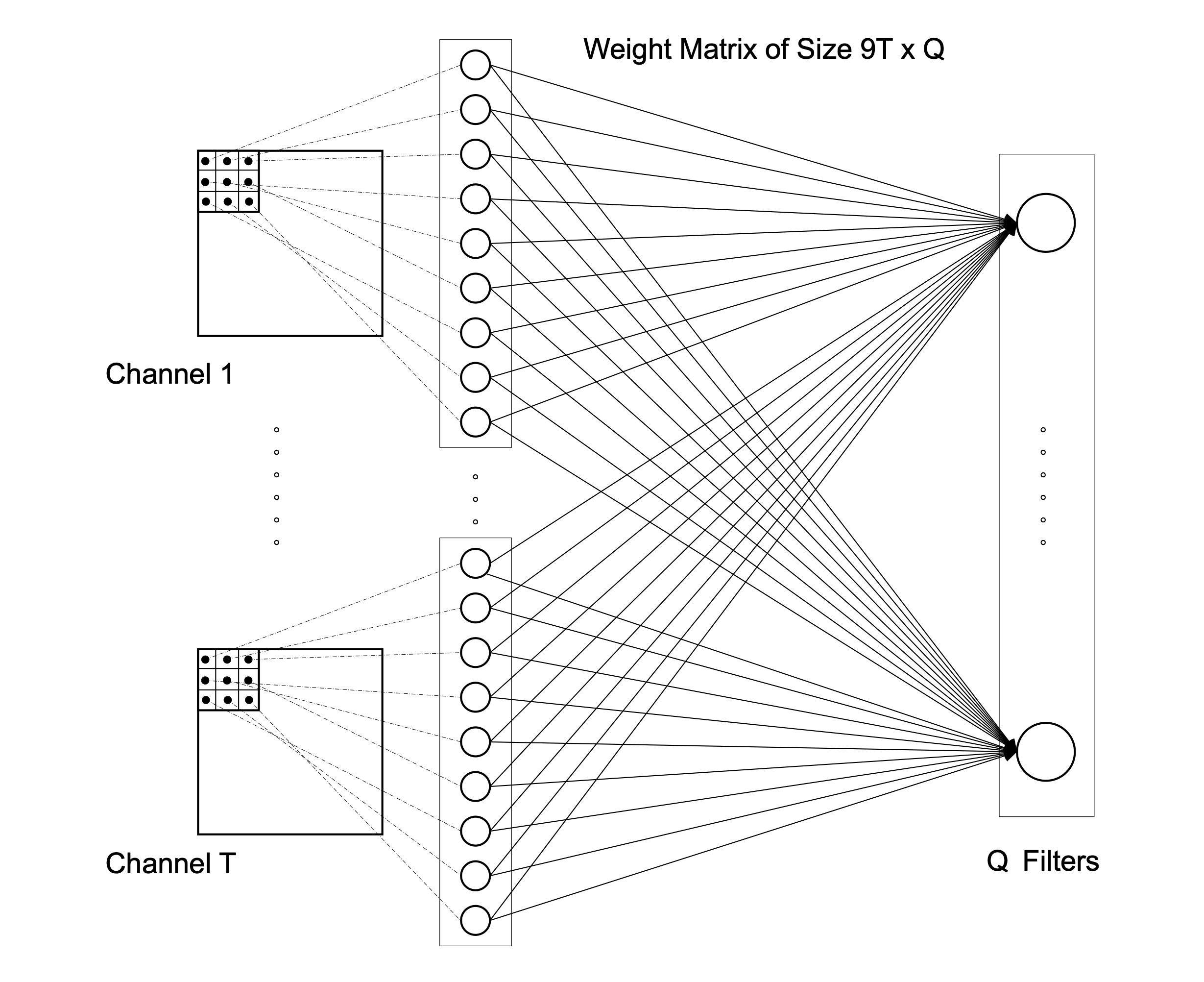}
	\caption{Illustrative framework of SteinGLM initialization for a convolutional layer.}
	\label{cnn_demo}
\end{figure}

The convolution operation is performing an element-wise multiplication between each image patch and filters. Assume the filter is of size $l_{h} \times l_{w}$,
then a convolution operation can be viewed as a matrix multiplication between a flattened image patch ($1 \times l_{h}l_{w}$) and a flattened filter ($l_{h}l_{w} \times 1$). An illustrative framework for using SteinGLM to initialize CNNs is shown in Figure~\ref{cnn_demo} (biases are ignored for simplicity). The convolution operation can be transformed into a typical fully-connected network: each patch in an input image corresponds to a new input sample (e.g., with $3 \times 3 = 9$ features) and each filter corresponds to the connecting weights of a hidden node. For a $T$-channel image to be convoluted by $Q$ filters, the unknown weight matrix is of size $T l_h l_w \times Q$, which can be initialized by Stein's method. Therefore, CNNs can also be initialized using the SteinGLM method introduced in this paper. Note the for each convolutional layer to be initialized, the size of new samples is $nP$ (where $P$ is the number of patches per image), hence additional computing resource is required for deriving the second-order score functions.

Using this transformation, a demo experiment on the MNIST dataset is conducted. 
To demonstrate the binary classification tasks, we pick 4 pairs of digits that are considered relatively hard to discriminate, including (3, 5), (4, 9), (7, 9) and (5, 8). We consider a multi-layer CNN with 9 convolutional layers, 3 pooling layers, and 1 fully-connected layer. The convolutional and pooling layers are divided into three blocks, where each block has 3 convolutional layers (with filter size $3 \times 3$, and ``SAME'' padding) followed by a $2 \times 2$ max-pooling layer. The numbers of filters in the first, second, and third blocks are set to be 9, 18, and 36, respectively. A fully-connected layer with 20 nodes is used before the output layer. 

During training, the batch size is set to 128, the number of training epoch is set to 20, and an $\ell_{2}$-regularization is performed for all weight matrices. When using SteinGLM, all the settings are consistent with previous experiments, except that the bias terms are ignored for the convolutional layers. The Stein's method is conducted 9 times for the convolutional layers and 1 time for the fully-connected layer (each with $n$ samples). The final layer is initialized using the GLM method with $n$ samples.

\begin{figure}[!t]
	\centering
	\includegraphics[width=1.0\textwidth, height=0.66\textwidth]{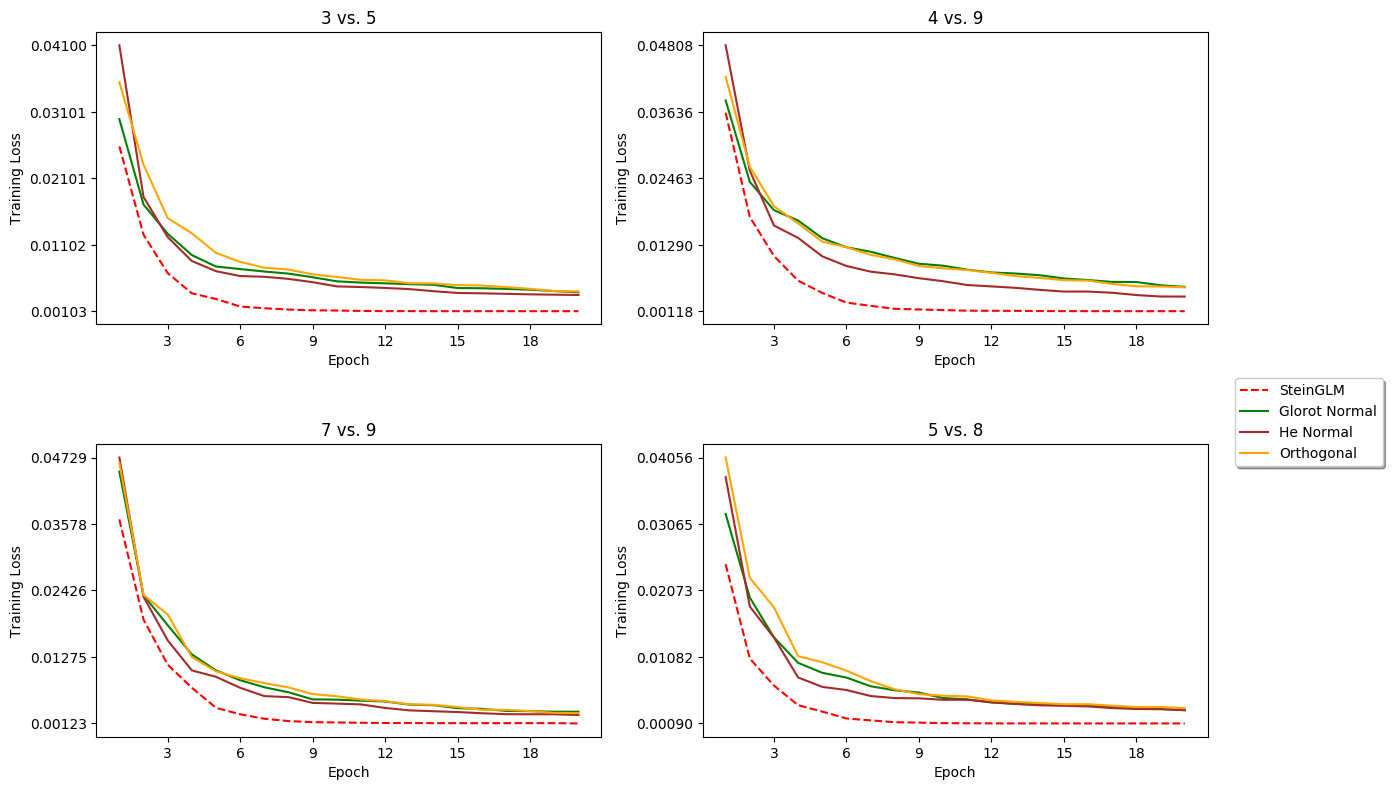}
	\caption{The trajectories of training losses for Mnist dataset.}
	\label{cnn_mnist}
\end{figure}

Figure~\ref{cnn_mnist} reports the training loss curves of different initialization methods. It can be observed that the proposed SteinGLM algorithm converges much faster in each binary classification task. Meanwhile, the validation and test set performances do not show significant difference among the compared methods. From such demonstrative experimental result, we find that the proposed SteinGLM method can speed up the convolutional neural network training, and its full potential will be explored in the future work.

\section{Conclusion} \label{Conclusion}
In this paper, we propose an effective and efficient method for training multi-layer feedforward neural networks. The key novelty lies in a new initialization scheme based on the second-order Stein's identity and generalized linear modeling. The numerical experiments show its superior performance in both computing time and accuracy against other benchmark methods. Some potential directions are worth further investigation. First, the input variables are currently assumed to follow the normal distribution. It is of our interest to develop a fast algorithm to dynamically learns the score function of general inputs. Second, only the sigmoid and hyperbolic tangent activation functions are considered in the present paper. We plan to take into account other activations (e.g. ReLU) as well. Third, as we have briefly demonstrated in Section~3.4, SteinGLM may help to speed up the training of convolutional neural networks, while its generalization performance does not outperform as well as in standard multi-layer neural networks.  
It is our plan to further investigate a more effective strategy for initializing convolutional neural networks.

\clearpage
\bibliographystyle{plainnat}
\bibliography{steinglm}
	
\end{document}